\documentclass[10pt,twocolumn,letterpaper]{article}

\usepackage{cvpr}
\usepackage{times}
\usepackage{epsfig}
\usepackage{graphicx}
\usepackage{amsmath}
\usepackage{amssymb}
\usepackage{algorithm}
\usepackage{algorithmic}

\usepackage[pagebackref=true,breaklinks=true,letterpaper=true,colorlinks,bookmarks=false]{hyperref}

 \cvprfinalcopy 

\ifcvprfinal\fi
\begin{document}

\title{Deep Multi-Spectral Registration Using Invariant Descriptor Learning}

\author{Nati Ofir\\
\and
Shai Silberstein\\
\and
Hila Levi\\
\and
Dani Rozenbaum \and
Yosi Keller \and
Sharon Duvdevani Bar
}

\maketitle

\begin{abstract}
	In this paper, we introduce a novel deep-learning method to align cross-spectral images. Our approach relies on a learned descriptor which is invariant to different spectra. Multi-modal images of the same scene capture different signals and therefore their registration is challenging and it is not solved by classic approaches. To that end, we developed a feature-based approach that solves the visible (VIS) to Near-Infra-Red (NIR) registration problem. Our algorithm detects corners by Harris \cite{harris} and matches them by a patch-metric learned on top of a network trained on CIFAR-10 \cite{krizhevsky2009learning} dataset. As our experiments demonstrate we achieve a high-quality alignment of cross-spectral images with a sub-pixel accuracy. Comparing to other existing methods, our approach is more accurate in the task of VIS to NIR registration.
\end{abstract}

\section{Introduction}

\begin{figure}
	\centering
	\includegraphics[width=120px]{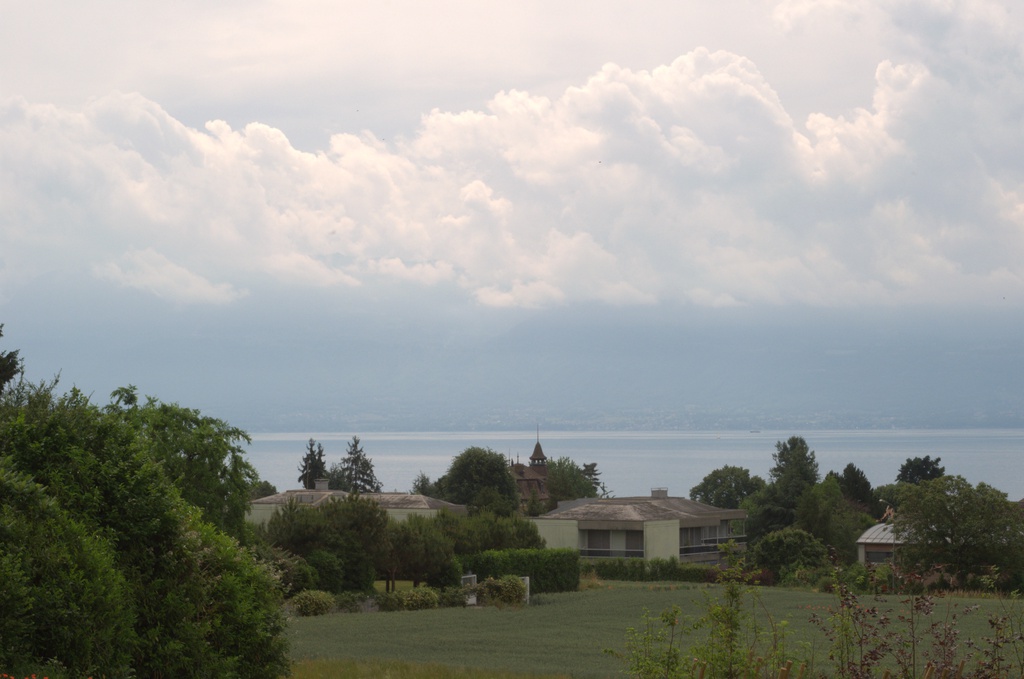}~
	\includegraphics[width=120px]{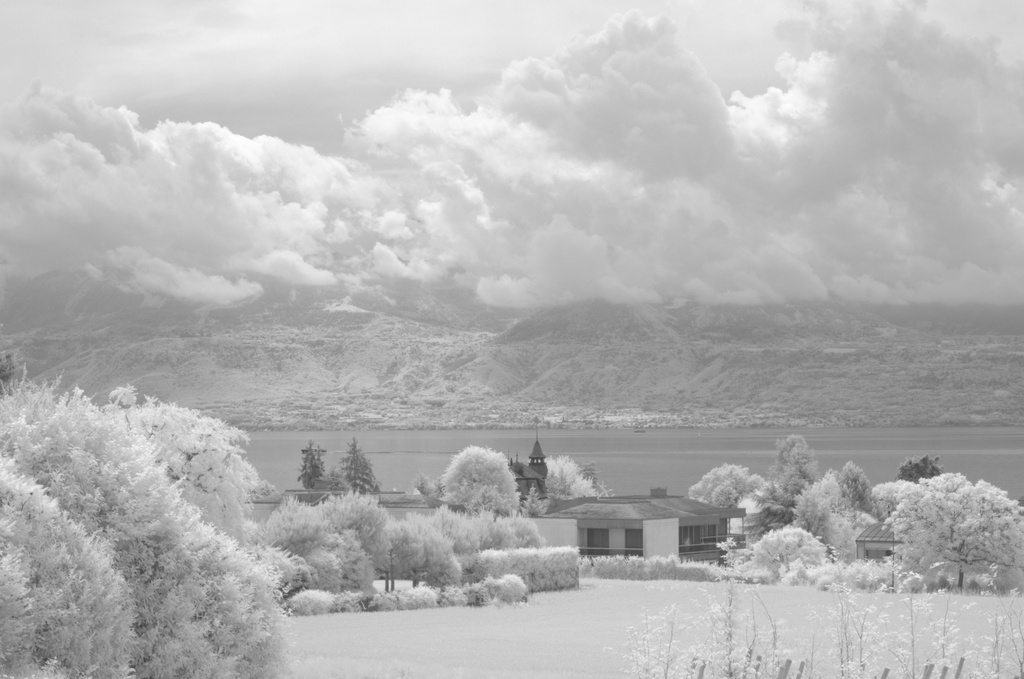}
	
	\caption{Example of a pair of multi-spectral images. Left: RGB image of the VIS channel $0.4-0.7\mu m$, contains the information of the color. Right: gray scale NIR image $0.7-2.5\mu m$ captures fine details of the far mountains.}
	\label{fig:1}       
\end{figure}

This paper addresses the problem of multi-spectral registration, and is aimed specifically to the visible (VIS) $0.4-0.7 \mu m$ and Near-Infra-Red (NIR) $0.7-2.5\mu m$ channels. Different spectra capture different scenes, therefore their registration is challenging and cannot be solved by state-of-the-art approaches for geometric alignment \cite{zitovaSurvey}. See Figure \ref{fig:1} for example, the VIS channel captures the color of the scene while the NIR channel captures more details about the far objects. These two images are different by their nature. In this work we introduce a method for performing registration of this kind of images based on deep learning.

Our approach is based on metric learning of cross-spectral image patches. In the first stage we detect feature points by Harris \cite{harris} corner detector. In the second stage, we match them to derive the global transformation between the input images. Since the patches around these points are different by their nature, SIFT \cite{SIFT} matching will not produce correct results. Therefore, we match them by deep-learning approach as follows. Our network is trained on CIFAR-10 \cite{krizhevsky2009learning} dataset and is geared to classify $32\times 32\times 3$ patches of the RGB visible channel into 10 classes. By removing its last soft-max layer and achieve an informative $64\times 1$ descriptor for each such RGB patch. Now, we train this trimmed net from scratch on NIR patches, such that cross spectral patches of the same object are trained to produce the same descriptor. Then we get two nets with the same architecture and different weights, the first for VIS descriptor and the second for NIR. These two networks induce a metric between cross spectral patches, which is the Euclidean distance of the two descriptors. As we show experimentally, this metric is an accurate basis for classification of multi-spectral patches to same or different, and therefore it is also a basis for our feature-based registration.

The paper is organized as follows. In Section \ref{sec:previous} we cover previous work on the topic of multi-modal registration. In Section \ref{sec:descriptor} we introduce our learning scheme of a deep-descriptor invariant to different wavelengths on top of a network trained on CIFAR-10 dataset. Then, in Section \ref{sec:registration}, we explain how to use this descriptor to perform multi-spectral registration. In Section \ref{sec:experimetns} we demonstrate evaluations of the accuracy of our registration algorithm compared to other existing approaches.

\section{Previous Work} \label{sec:previous}

Image registration is an important task of computer vision which has many related works. Image registration methods \cite{survey,zitovaSurvey} are the basis for many applications such as image fusion and 3d object recovery. Early methods rely on basic approaches such as solving translation by correlation \cite{correlation}. \cite{lucas1981iterative} solves registration based on the gradients of the image. \cite{reddy1996fft} solves global registration using a FFT-based method. Advanced methods are using key-points \cite{harris,SIFT} and invariant descriptors to find the geometric alignment \cite{SIFT_Registration}.

Multi-modal images registration is also addressed by several works over the last decades \cite{chen2015sirf,shen2014multi}. In \cite{mutualInformaion} registration was carried out based on maximization of mutual information. \cite{multiSpectralSIFT,irani1998robust,keller2006multisensor,li1995contour} offer to utilize contours and gradients for registration. If solving only displacement, cross spectral registration can be executed by measuring correlation on Sobel image \cite{sobel}, or Canny image \cite{Canny}. A group of works was focused on the specific task of visible to Near-Infrared (NIR) registration \cite{multiSpectralSIFT,keller2006multisensor}.  \cite{lghd2015} solves registration by FAST features \cite{takacs2010unified} and unique descriptors for non linear intensity variations. \cite{aguilera2016learning} was the first to measure cross-spectral similarity by Convolutional-Neural-Network (CNN). Their approach indeed manages to classify pairs of multi-spectral patches into same or different, but unfortunately, do not induce a metric. Therefore, if two patch-pairs are found similar by their network, it cannot be derived which one is more similar than the other. This knowledge is crucial for finding the best match for a feature-point. Our approach instead utilizes CNN's to measure the distance between cross spectral patches, and also to produce a continuous score of how similar they are. This measure forms a solid basis for multi-spectral registration as described in Sections \ref{sec:descriptor} and \ref{sec:registration}.

\section{Multi-Spectral Descriptor Learning} \label{sec:descriptor}

We propose to align a pair of cross spectral images by a feature based approach. In order to find the global transformation between the feature points, we need a mechanism to match them. In this section we introduce an approach to match feature points from different spectral channels by their deep-descriptor.

Given a VIS channel patch $P_v$ and a NIR patch $P_n$ we offer to learn a metric that measures the similarity distance between them. The descriptor of $P_v$ is computed from the trimmed network trained on CIFAR-10 \cite{krizhevsky2009learning}. Figure \ref{fig:1} summarizes this network architecture. Denoting this network by $Net_v$, then the descriptor of the visible channel is $D_v = Net_v(P_v)$. Now, we would like to learn a network $Net_n$ for the NIR channel, with the same architecture as in Figure \ref{fig:1} but with different weights. The NIR descriptor is $D_n = Net_n(P_n)$. We seek for weights of $Net_n$ such that the descriptor $D_n$ would be invariant to different wavelengths. It means that the distance of corresponding patches, $P_v$ and $P_n$

\begin{equation} \label{eq:distance}
distance(P_v,P_n) = || Net_v(P_v)-Net_n(P_n)||_2^2
\end{equation}

would be significantly less than the distance of non-corresponding patches.

\begin{figure}
	\centering
	\includegraphics[width=120px]{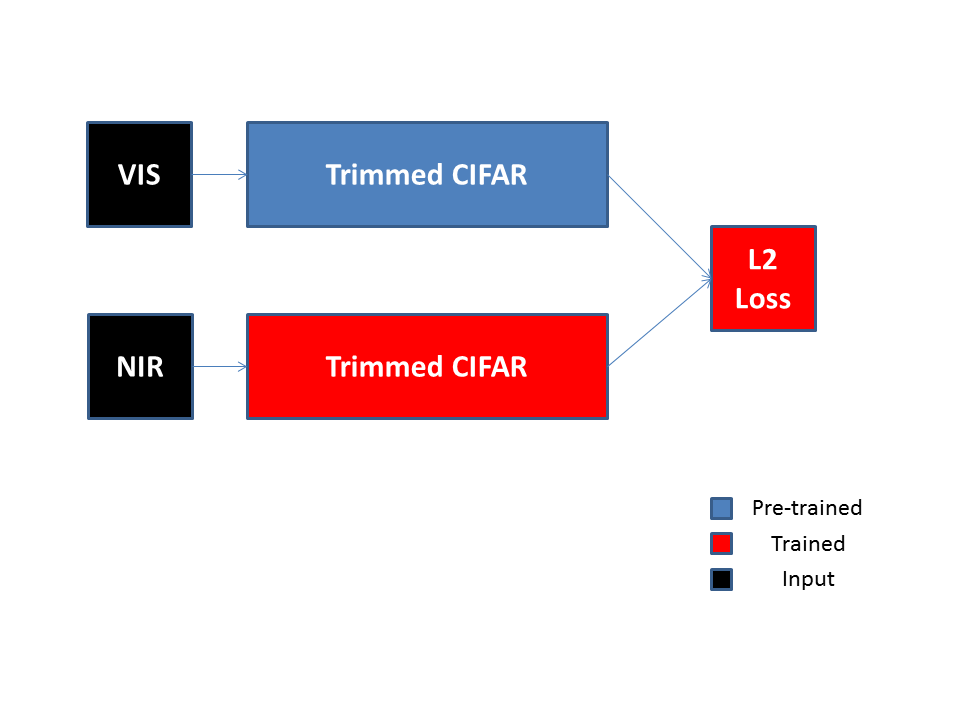}~
	\includegraphics[width=120px]{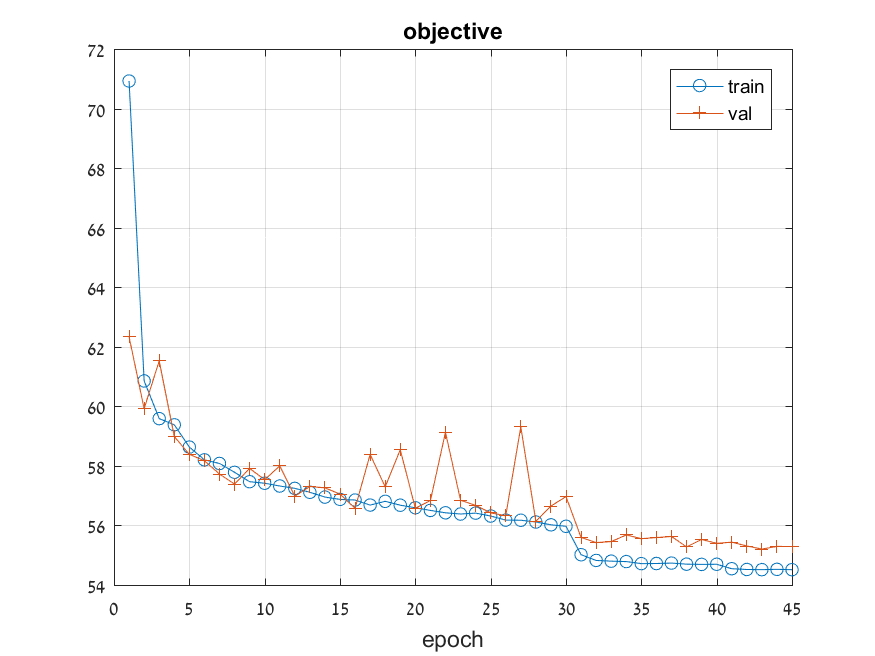}~
	
	\caption{Left: trained network architecture of our metric learning scheme. Right: training process of our NIR descriptor network. It can be seen that the L2 loss decreases over the epochs, while the validation objective is slightly above the training graph.}
	\label{fig:2}       
\end{figure}

We learn the weights of $Net_n$ as follows. We use the dataset of \cite{multiSpectralSIFT} that contains over 900 aligned images from the VIS and NIR channels. For every image we apply Harris corner detector \cite{harris} and extract around 1000 patch center. By that process we store over 100,000 corresponding pairs of cross-spectral patches, each such pair being a training example. The input for $Net_n$ is the NIR patch $P_n$, while the label is $D_v = Net_v(P_v)$. By that approach we teach the network to output the visible descriptor for a NIR input. This network is responsible for maintaining the invariance to spectral channels of our distance metric. Figure \ref{fig:2} demonstrates the convergence of our training process and the trained network architecture. It can be seen that the L2 distance is decreasing over the epochs, and that the validation graph is closed to the training graph, indicating that there is no over-fitting.

An alternative for computing the spectral distance in equation \eqref{eq:distance} is to measure similarity would be to use a 2-channel network as described in \cite{aguilera2016learning}. This 2-channel network would receive the two patches $P_v$ and $P_n$ as an input and outputs 1 if they are same and -1 if they are different. It is trained by pairs of positive and negative multi-spectral patches. Unfortunately, this architecture cannot be used for registration because it does not form a metric. If we found two matches for a patch, it is necessary to know which one is more similar, and this cannot be deduced in a 2-channel network that acts as binary classifier. On the contrary, our metric indicates a distance between the patches, and therefore supply to the algorithm a boolean indicates if the patches are same, but also a similarity score. Additional problem of the 2-channel architecture is its run-time. This network requires to run a forward pass for each pair of corners, and it is an expensive procedure that can take minutes. In our architecture, a forward pass is calculated for each corner separately, and only L2 distances are computed for every pair. 

In Section \ref{sec:experimetns} we will show that our metric can be used to classify between same and different pairs of cross-spectral patches with high accuracy. In the next Section \ref{sec:registration} we explain how to use this metric to compute the multi-spectral registration.

\begin{table}
	\centering
	\begin{tabular}{ c | c | c | c | c | c}
		Layer & Type & Output Dim & Kernel & Stride & Pad\\
		\hline
		1 & convolution & 32 & 5$\times$5 & 1 & 2\\
		2 & max-pooling & 32 & 3$\times$3 & 2 & 0\\
		3 & ReLU & 32 & - & 1 & 0 \\
		4 & convolution & 32 & 5$\times$5 & 1 & 2\\
		5 & ReLU & 32 & - & 1 & 0\\
		6 & avg-pooling & 32 & 3$\times$3 & 2 & 0\\
		7 & convolution & 64 & 5$\times$5 & 1 & 2\\
		8 & ReLU & 64 & - & 1 & 0\\
		9 & avg-pooling & 64 & 3$\times$3 & 2 & 0\\
		10 & convolution & 64 & 4$\times$4 & 1 & 0
	\end{tabular}
	\caption{Architecture of the trimmed network trained on CIFAR-10 dataset. This net gets an $32\times 32$ image patch as an input and outputs its spectral-invariant descriptor.}
	\label{table:1}
\end{table}

\section{Multi-Spectral Image Registration} \label{sec:registration}

We use the metric of cross-spectral patches described in Section \ref{sec:descriptor} to form a deep feature based registration. Our approach consists of three stages: corner detection by Harris \cite{harris}, corners matching by our deep descriptor and finally computation of the global geometric transformation by Random-Sample-Consensus (RANSAC) \cite{ransac} mechanism for outliers rejection.

\textbf{Corners Detection.}
Denote by $V$ the VIS channel image and by $N$ the corresponding NIR image. We use the method described in \cite{harris} to extract the corresponding group of corners $C_v$ and $C_n$. Each such corner is a local maximum in the Harris score image:
\begin{equation}
S = \lambda_1\lambda_2-k(\lambda_1+\lambda_2)^2 = \det(A)-k\cdot trace^2(A),
\end{equation}
where $\lambda_1,\lambda_2$ are the eigenvectors of the matrix of derivatives for each pixel: 
\begin{equation} \label{eq:harrisMatrix}
A = \sum_u \sum_v w(u,v)
\begin{pmatrix}
I_x(u,v)^2 & I_x(u,v)I_y(u,v) \\ 
I_x(u,v)I_y(u,v) & I_x(u,v)^2
\end{pmatrix}.
\end{equation}
$I_x,I_y$ are the horizontal and vertical derivatives of the input image respectively. Since this corner detection method is based on local gradients, it is relatively invariant to multi-spectral images and therefore the group of corners $C_v$ and $C_n$ has a large overlap. This characteristic of the feature extraction is necessary for the success of our whole scheme.

\textbf{Feature Matching.}
We want to match the points in $C_v$ to those in $C_n$ to form the group of all matches $M$. For every point $p_v \in C_v$ we find the best match $p^*_n\in C_n$ as follows. Firstly, we compute the descriptor of $C_v$ by $Net_v$ and the descriptors of $C_n$ by $Net_n$. The complexity of all this forward passes is $|C_v|+|C_n|$ which is practical and feasible for real time registration. If we used a 2-channel network for computing similarities, the complexity would be $|C_v|\times|C_n|$ and the run-time would be huge. A match $m \in M$ is a pair $(p_v,p^*_n)$, such that $p^*_n$ is the nearest neighbour of $p_v$ according to our deep metric. It means that their descriptors $D_v$ and $D_{p^*_n}$ are the closest out of all other possible matches.

\textbf{Transformation Computation.}
Now, we use the group of all matches $M$ to form the final global transformation $H$ between the images $V$ and $N$. Typically, most of the matches in $M$ are outliers. Therefore, we compute the transformation, $H$ by RANSAC \cite{ransac} outliers rejection method. The transformation $H$ for every sample of matches from $M$ is computed by least-squares. We look for the largest sub-group of $M$ that accepts on the same transformation $H$, this group contains the inlier matches. We can restrict $H$ to be affine, rigid or translation only. As the degree of freedom of $H$ is getting lower, the accuracy of our registration becomes higher. The transformation found by RANSAC is the final result of our algorithm, the score of success is the ratio of inliers divided by $|M|$. A score greater than $0.1$ indicates a successful run of our method. In Section \ref{sec:experimetns} we evaluate the accuracy of our approach.

\section{Experiments} \label{sec:experimetns}

\begin{figure}
	\centering
	\includegraphics[width=60px]{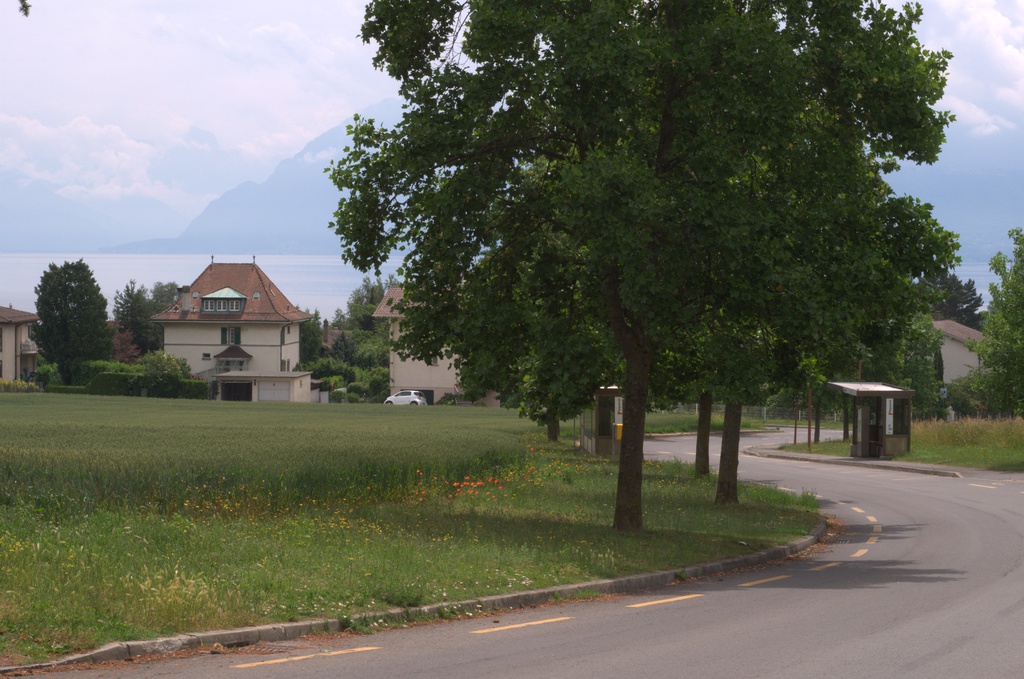}~
	\includegraphics[width=60px]{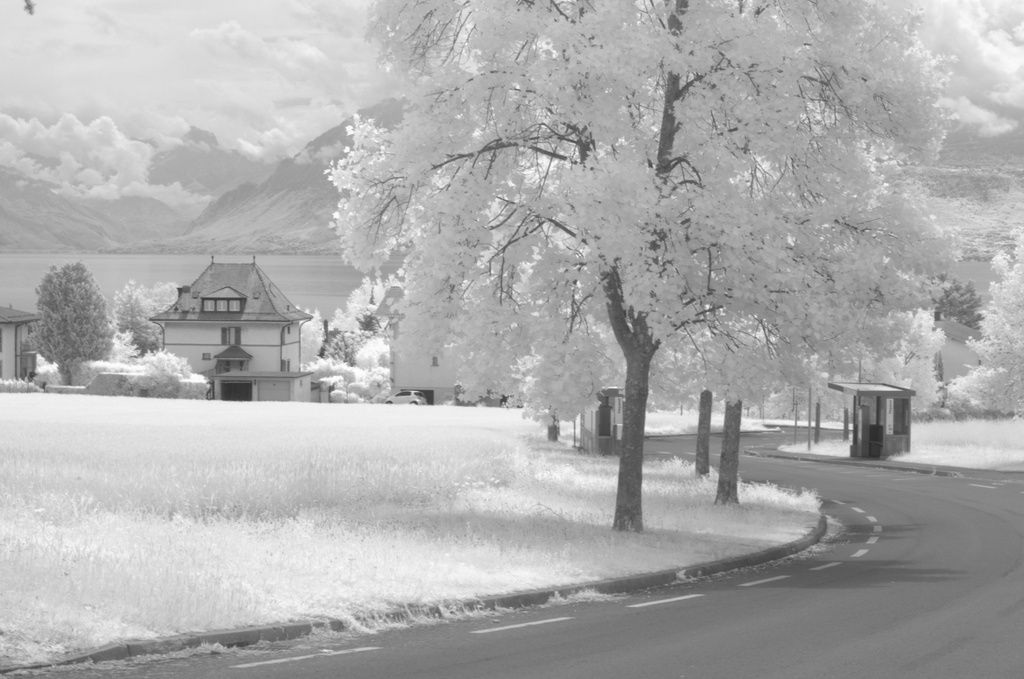}\\[0.1cm]
	\includegraphics[width=60px]{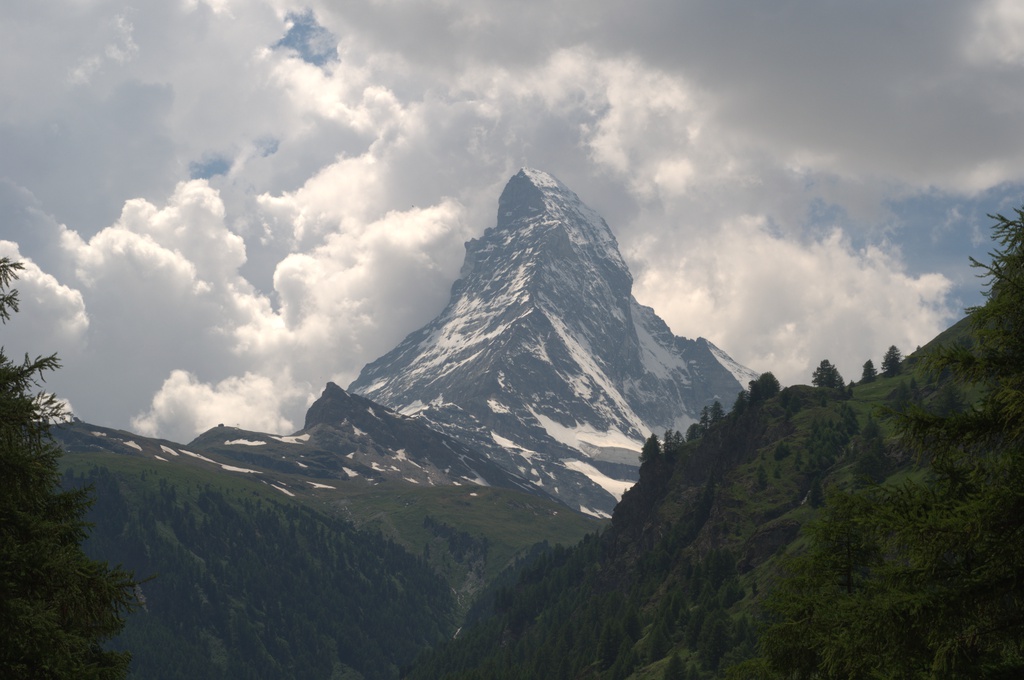}~
	\includegraphics[width=60px]{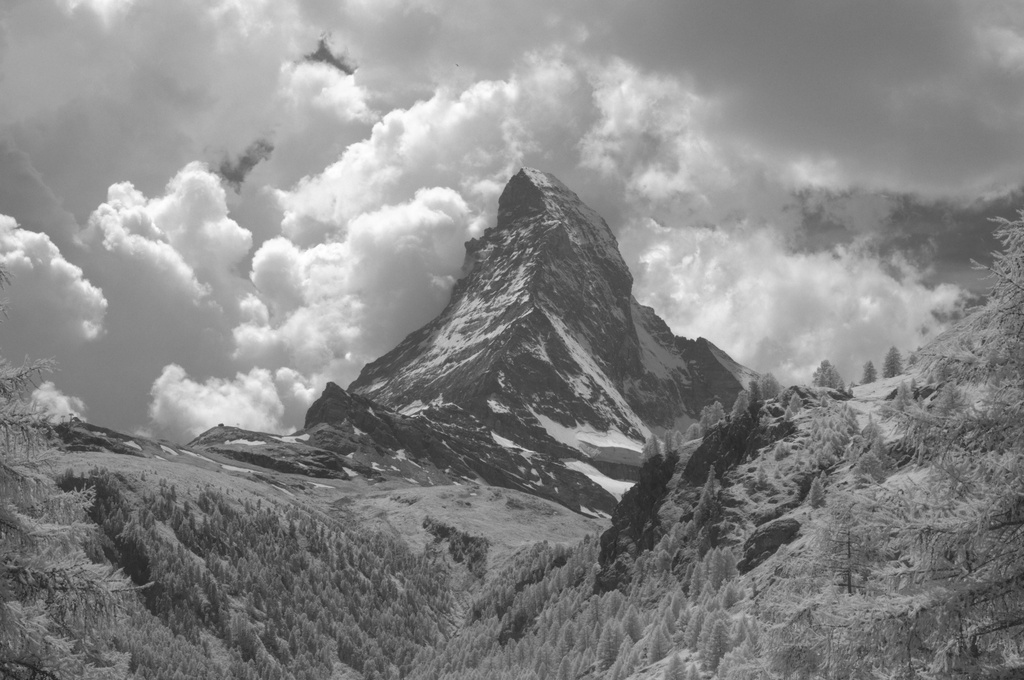}
	
	\caption{Pairs of aligned cross-spectral images from the dataset we used to train and evaluate our method \cite{multiSpectralSIFT}.}
	\label{fig:6}       
\end{figure}

We trained and tested our method on cross-spectral images from the datasets of \cite{multiSpectralSIFT}. This dataset contains over 900 aligned images from the VIS and NIR channels. In Figure \ref{fig:6} we show examples of pair of images from this dataset. Our code is implemented in Matlab using MatConvNet library \cite{Vedaldi15}. The runtime of our registration is around 10 seconds per pair of images and can be further reduced by utilization of GPU and parallel computing. The training time for our network is one hour on a Titan-X GPU. We trained the network with learning rate of 0.005 and weight decay of 0.0004. For evaluation of our registration accuracy we manually simulated transformations on the dataset of aligned cross-spectral images and tried to recover them automatically by our approach. For each run of a simulation we recored the error which is the Euclidean distance of a specific parameter, for example the error between a simulated translation to the one found by our code. We compared our method to several different approaches for multi-spectral registration. The first approach is to use edge descriptors and to match them by binary correlation, this is still feature based and can solve any type of transformation. Additional approaches solve only translation, among them correlation of Canny \cite{Canny} images, correlation of Sobel \cite{sobel} or maximization of mutual information. We also compare to the feature based approach of LGHD descriptor \cite{lghd2015}.

\begin{figure}
	\centering
	\includegraphics[width=100px]{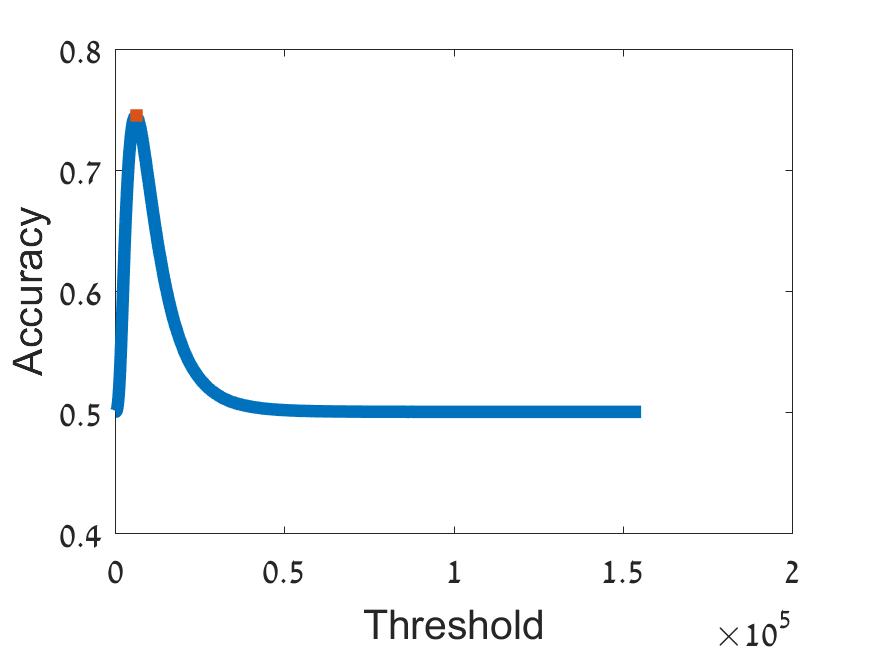}~
	\includegraphics[width=100px]{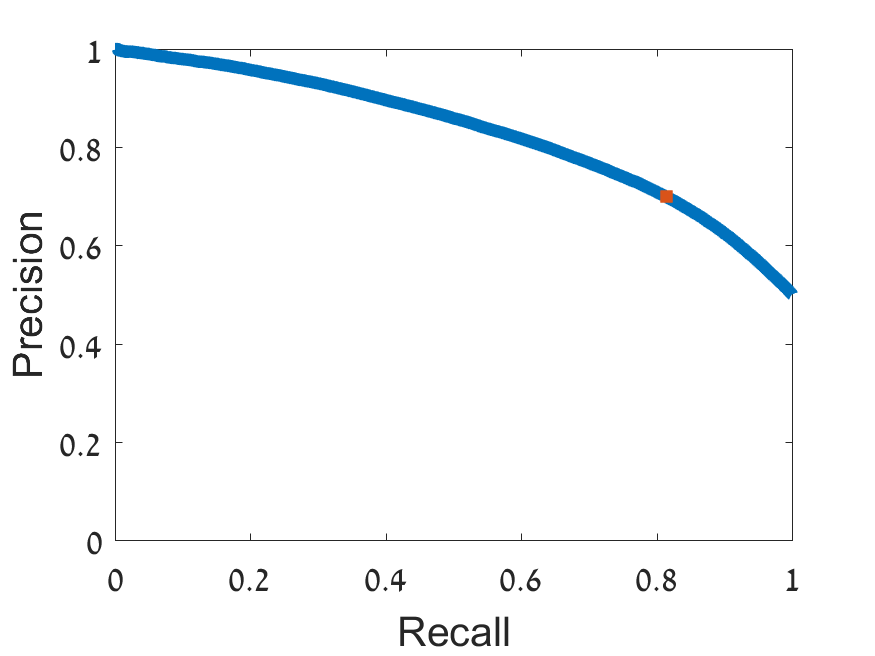}	
	\caption{Evaluation of our deep-metric as a binary classifier to same or different pairs of cross-spectral patches. Left: accuracy of the classifier as a function of threshold, the top score is $0.74$. Right: precision-recall graph of the classifier, the obtained F-score is $0.75$.}
	\label{fig:3}       
\end{figure}

In Figure \ref{fig:3} we show our results when trying to classify pair of patches to similar or different. The positive set is the pair of patches around corners in the dataset while the negative examples are produced by random matching. The accuracy of our binary-classifier is $0.74$ when selecting to correct threshold on the L2 distance between the descriptors. The F-measure \cite{powers2011evaluation}, $F = \frac{2*precision*recall}{precision+recall}$, is 0.75 and it is achieved with a similar threshold for maximizing the accuracy.

\begin{figure}
	\centering
	\includegraphics[width=160px]{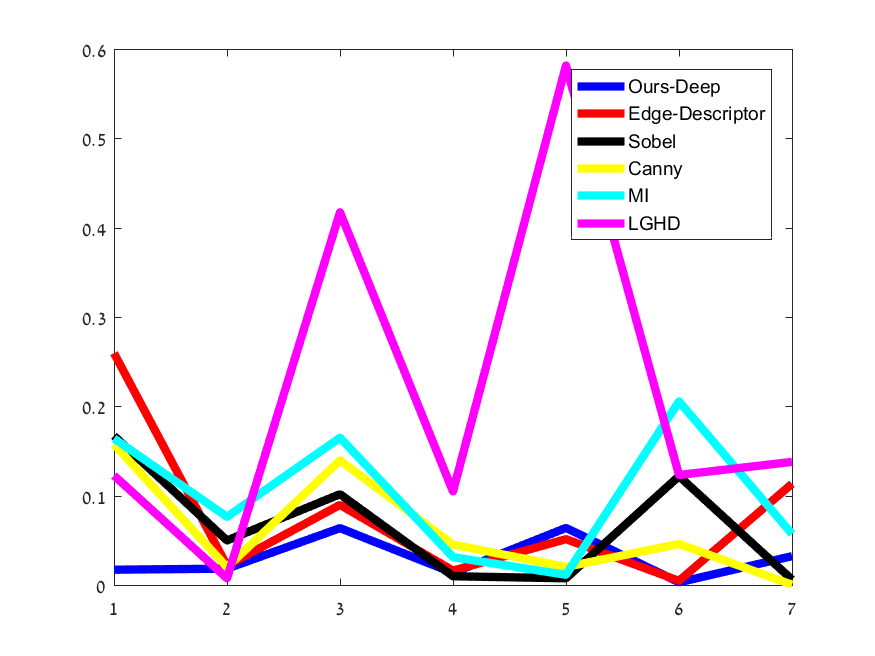}
	
	\caption{Error in pixels of cross-spectral registration methods over sample of scenes. As can be seen, our deep method achieves the lowest error in most of the examples.}
	\label{fig:4}       
\end{figure}

Table \ref{table:2} compares the different methods when trying to solve translation only. It can be noticed that our deep method achieves the lowest error which is very close to 0 pixels. In Figure \ref{fig:4} we plot this error along sample of different scenes. It can be seen that our error is the lowest across almost all the scenes.

\begin{figure}
	\centering
	\includegraphics[width=100px]{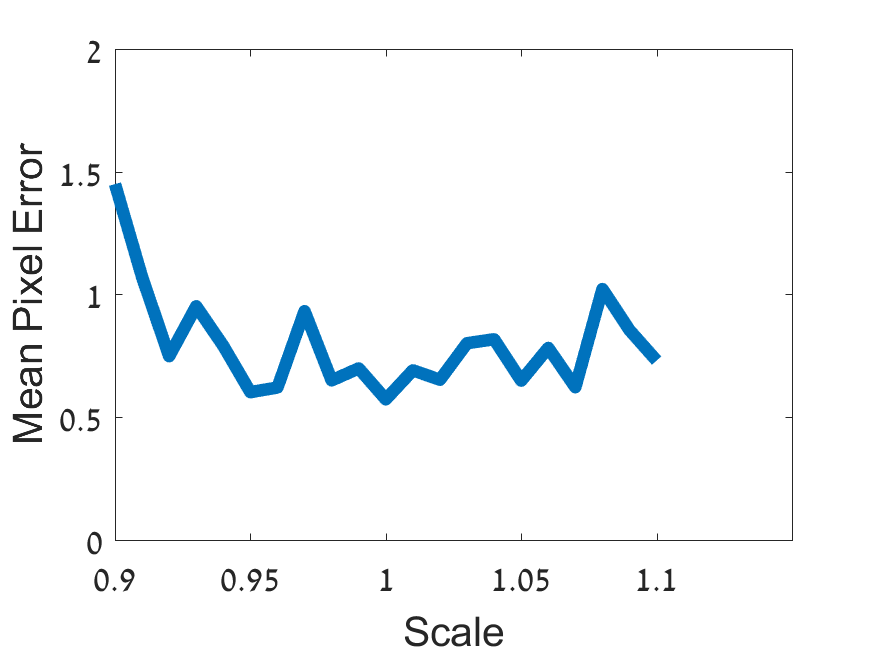}~
	\includegraphics[width=100px]{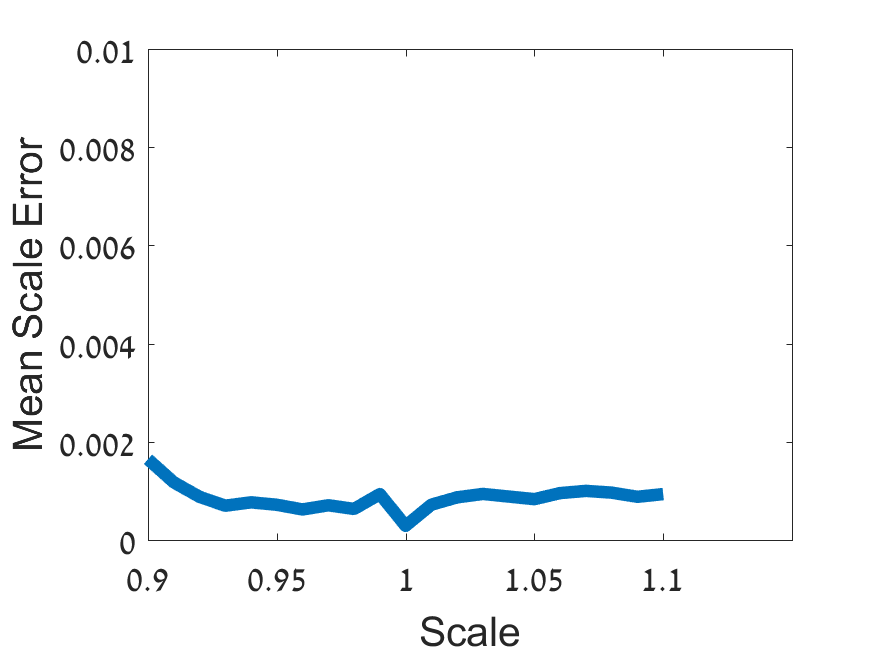}	
	\caption{Evaluation of registration error across simulated scaling transformations. Left: error of the translation parameters when solving scales from 0.9 ro 1.1. Right: error of the scaling parameter acrros the same range of scalings between the cross-spectral images. The translation error is around 1 pixels while the scaling error is negligible.}
	\label{fig:5}       
\end{figure}

Figure \ref{fig:5} shows the error of our deep-method across different scaling of the simulated transformation. When solving scaling we achieve an error of around one pixel in all scalings levels. In the difference of the scaling parameter we gain a negligible error below 0.002. Overall, our multi-spectral registration is accurate and solves complex transformations. 

\section{Conclusions} \label{sec:conclusions}

We introduced a novel method for multi-spectral registration that utilized an invariant deep descriptor of cross-spectral patches. For that end, we trained a network to extract such descriptor for NIR patches. This network together with the trimmed network pre-trained on CIFAR-10 for RGB patches, form a metric between multi-spectral patches. Our experiments demonstrate that our metric-learning scheme is useful for classifying pair of patches to same or different. Moreover, it forms a basis for an accurate multi-spectral registration. In future work we plan to build and train a fully end-to-end network that will carry out all the stages of our feature based registration including corner detection and feature matching. In addition, we plan to train a generative adversarial networks to create a VIS image out of NIR image. 

\begin{table}
	\centering
	\begin{tabular}{ l | c }
		Algorithm & VIS-NIR \\
		\hline
		Our method & \textbf{0.03} \\
		Edge-Descriptor & 0.08 \\
		Canny & 0.07 \\
		Sobel & 0.07 \\
		Mutual Information & 0.11 \\
		LGHD & 0.21
	\end{tabular}
	\caption{Error in pixels of multi-spectral registration when searching for translation only. Our deep method is compared to edge descriptors approach, correlation of Canny, correlation of Sobel, maximization of mutual-information and LGHD. As can be seen (in bold), our deep-algorithm achieves the highest accuracy.}
	\label{table:2}
\end{table}

{\small
	\bibliographystyle{ieee}
	\bibliography{../egbib}
}

\end{document}